%% file: main.tex
\documentclass[runningheads]{llncs}
\usepackage[T1]{fontenc}
\usepackage{amsmath}
\DeclareMathOperator*{\argmin}{arg\,min}
\usepackage{amssymb}
\usepackage{algorithm}
\usepackage{algpseudocode}
\usepackage{graphicx}
\usepackage{enumerate}
\usepackage{enumitem}
\usepackage{wrapfig}
\usepackage{bm}
\usepackage{tabularx}
\usepackage{booktabs}
\begin{document}
\title{Enhancing RL Generalizability in Robotics through SHAP Analysis of Algorithms and Hyperparameters}
\titlerunning{SHAP for RL Generalizability in Robotics}
%
\author{Lingxiao Kong\inst{1,2}\orcidID{0009-0003-1968-7025} \and
Cong Yang\inst{3}\orcidID{0000-0002-8314-0935} \and
Oya Deniz Beyan\inst{1,2,4}\orcidID{0000-0001-7611-3501} \and
Zeyd Boukhers\inst{1,4}\orcidID{0000-0001-9778-9164}}
\authorrunning{L. Kong et al.}
%
\institute{Fraunhofer Institute for Applied Information Technology FIT, Germany \\
\email{\{lingxiao.kong,oya.deniz.beyan,zeyd.boukhers\}@fit.fraunhofer.de} \and
University of Cologne, Germany \and
Soochow University, China \\
\email{cong.yang@suda.edu.cn} \and
University Hospital of Cologne, Germany}
\maketitle              
\begin{abstract}
Despite significant advances in Reinforcement Learning (RL), model performance remains highly sensitive to algorithm and hyperparameter configurations, while generalization gaps across environments complicate real-world deployment. Although prior work has studied RL generalization, the relative contribution of specific configurations to the generalization gap has not been quantitatively decomposed and systematically leveraged for configuration selection. To address this limitation, we propose an explainable framework that evaluates RL performance across robotic environments using SHapley Additive exPlanations (SHAP) to quantify configuration impacts. We establish a theoretical foundation connecting Shapley values to generalizability, empirically analyze configuration impact patterns, and introduce SHAP-guided configuration selection to enhance generalization. Our results reveal distinct patterns across algorithms and hyperparameters, with consistent configuration impacts across diverse tasks and environments. By applying these insights to configuration selection, we achieve improved RL generalizability and provide actionable guidance for practitioners.

\end{abstract}

\input{sections/1_introduction}
\input{sections/2_related_work}

\input{sections/3_methodology}
\input{sections/4_experiments}
\input{sections/5_results}
\input{sections/6_conclusion}

\section*{Acknowledgements}
Funded by the European Union. This work has received funding from the European High Performance Computing Joint Undertaking (JU) and from the German Federal Ministry of Research, Technology and Space (BMFTR), the Ministry of Culture and Science of North Rhine-Westphalia (MKW NRW) and the Hessian Ministry of Science and Research, Arts and Culture (HMWK) under grant agreement No 101250682.

\bibliographystyle{splncs04}
\bibliography{mybibliography}

\end{document}

%% file: sections/1_introduction.tex
\section{Introduction}
Reinforcement Learning (RL) has emerged as a powerful paradigm due to its ability to learn through environmental interaction, with successful applications in game theory, robotics, and natural language processing~\cite{kong2025reinforcement}. However, RL models are typically built on lightweight policies (e.g., Multi-Layer Perceptron policies at the million-parameter scale), making them significantly smaller than modern Large Language Models (LLMs) and posing critical challenges for generalizability across diverse environments. This is particularly evident in Simulation-to-Simulation (Sim2Sim) and Simulation-to-Real (Sim2Real) transfers~\cite{hoefer2021sim2real}, where environmental discrepancies including physical constraints and failure costs necessitate training primarily in simulation, yet models frequently struggle to transfer effectively to new settings. Empirical studies in RL-Baselines3 Zoo~\cite{raffin2020rl-zoo3} demonstrate that even similar robotic tasks exhibit substantial performance variation across simulated environments.

RL model performance is highly sensitive to algorithm and hyperparameter configurations. For instance, learning rates significantly influence training dynamics: excessively high rates accelerate learning but may cause instability, while overly low rates result in slow convergence or entrapment in suboptimal local optima. Understanding how these configurations affect performance is crucial for improving RL generalizability and enabling successful Sim2Sim and Sim2Real transfers, as bridging this gap poses fundamental challenges to real-world deployment~\cite{zhao2020SRT,wagenmaker2024overcoming}. In this context, explainable AI has received growing attention for analyzing how different factors contribute to model behavior~\cite{milani2024explainable}, with SHapley Additive exPlanations (SHAP)~\cite{lundberg2017SHAP} quantifying each input's contribution by enumerating feature subsets and computing weighted marginal effects.

\begin{figure}[t]
   \begin{center}
\includegraphics[width=1.0\linewidth]{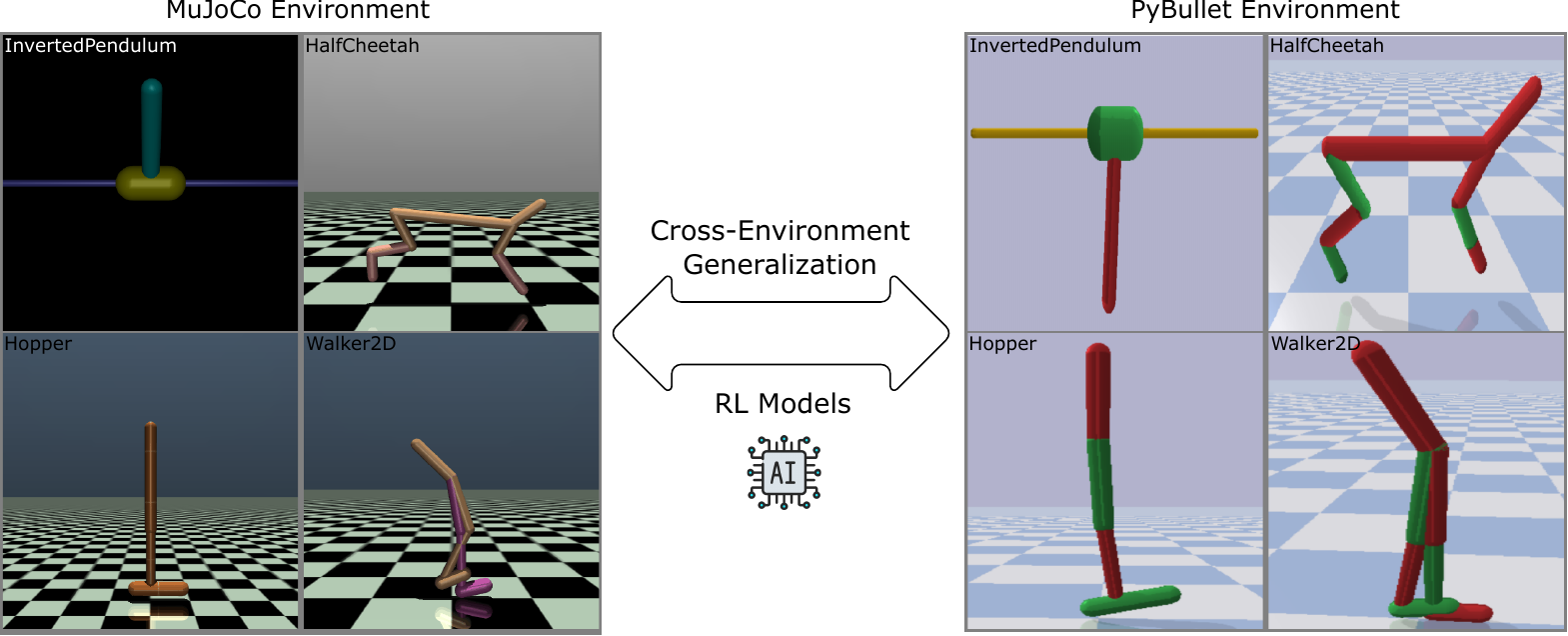}
   \end{center}
   \vspace{-0.2cm}
   \caption{Our work aims to analyze and guide cross-environment generalization of RL models through bidirectional transfer experiments: MuJoCo $\leftrightarrow$ PyBullet. }
   \vspace{-0.2cm}
   \label{fig:environments}
\end{figure}

To address \textbf{configuration pattern analysis} for transfer generalizability in RL, we propose a modular and reproducible framework integrating SHAP to analyze configuration impact patterns and guide selection strategies. We train numerous models with sampled configurations, evaluate them across training and testing environments, and employ SHAP to extract patterns that enable predictive modeling of optimal configurations while demonstrably improving generalizability. As illustrated in Figure~\ref{fig:environments}, we implement our framework on four pairs of standardized robotic tasks from Gymnasium~\cite{towers2024gymnasium} across MuJoCo and PyBullet physics engines bidirectionally, where the physics gap serves as a controlled proxy analogous to the Sim2Real gap. Our comprehensive analysis encompasses: main impacts of RL algorithms and hyperparameters; their interaction patterns; cross-task and cross-environment insights; and validated SHAP-guided configuration selection. To the best of our knowledge, this is the first work to leverage SHAP for explaining and guiding RL configuration patterns to improve generalization across robotic environments, with reproducible applicability to other configurations, tasks, and environments, providing practitioners with actionable insights into configuration impacts for improved generalizability.

Succinctly, our main contributions are: (1) establishing a theoretical foundation connecting generalizability with SHAP analysis; (2) presenting a SHAP-based framework for analyzing and guiding RL configuration impacts on generalizability; (3) developing an interface integrating robotic environments, RL agents, and explainable AI, with comprehensive experiments on four standardized robotic tasks; and (4) providing valuable insights and configuration selection for optimizing generalization. The code for our framework and experiments is available at \texttt{https://github.com/engineerkong/SHAP-RLROBO}.

%% file: sections/2_related_work.tex
\section{Background and Related Work}
In robotics, RL agents learn control policies by interacting with environments, updating policies based on observed states ($s$: e.g., joint angles, end-effector positions) and selecting actions ($a$: e.g., motor torques) via actuators. The environment yields new observations $s'$ and rewards $r$ determined by task objectives, driving iterative policy improvement across algorithms.

RL generalizability refers to the ability of trained RL agents to maintain effective performance when deployed in environments differing from their training conditions, as in Definition~\ref{def:generalizability}. RL-Baselines3 Zoo~\cite{raffin2020rl-zoo3} demonstrates significant performance variations across environments: for instance, A2C achieved a mean return of $2107.384$ in HalfCheetah-BulletEnv-v0 (PyBullet) compared to $3041.174$ in HalfCheetah-v3 (MuJoCo) under identical configurations, highlighting how environmental factors critically affect trained model performance~\cite{bian2024machine}.


\begin{definition}[Generalizability Problem]
\label{def:generalizability}
Given:
\begin{itemize}
    \item Source domain: $\mathcal{D}_S$ with physics parameters $\bm{\omega}_S$
    \item Target domain: $\mathcal{D}_T$ with physics parameters $\bm{\omega}_T$ where $\bm{\omega}_T \neq \bm{\omega}_S$
    \item Training configuration: $\bm{\theta} = (\mathcal{A}, h_1, \ldots, h_m)$ where $\mathcal{A}$ is the algorithm and $h_i$ are hyperparameters
\end{itemize}
The generalization gap is represented as $\Delta J(\bm{\theta})$ for policy $\pi_{\bm{\theta}}$ trained on $\mathcal{D}_S$:
\begin{equation}
\Delta J(\bm{\theta}) = J_S(\pi_{\bm{\theta}}) - J_T(\pi_{\bm{\theta}})
\end{equation}
\end{definition}

Recent studies highlight two critical directions of the generalizability challenge: environmental dynamics and model configuration~\cite{packer2018assessing,cobbe2019quantifying}. Different environments differ substantially in physics engines, computational architectures, and environmental parameters, creating significant generalization barriers~\cite{hwangbo2019learning}. Empirical evidence shows that hyperparameter tuning substantially affects model generalizability~\cite{katlav2025ai,smith2018disciplined}, with diverse approaches such as domain randomization, meta-learning, and transfer learning proposed to address it~\cite{katlav2025ai,pitkevich2024survey}, though these methods often sacrifice explainability~\cite{luo2024survey}.

To explain the effectiveness of models, the feature permutation method~\cite{adler2018auditing} evaluates feature significance by measuring prediction error changes when feature values are randomly shuffled, while surrogate models use interpretable models (e.g., decision trees) to approximate black-box model decisions~\cite{che2017interpretable}. Complementing these approaches, SHAP~\cite{lundberg2017SHAP} provides a game-theoretic framework treating each feature as a player and the prediction as the payoff, computing each feature's contribution through a surrogate explainer. As formalized in Definition~\ref{def:shapley}, Shapley values provide unified feature importance measures with desirable properties, enabling systematic analysis of configuration contributions to generalizability. 

\begin{definition}[Shapley Values for Generalizability]
\label{def:shapley}
Let $\mathcal{N} = \{0, 1, \ldots, m\}$ index configuration components (0 = algorithm, $1, \ldots, m$ = hyperparameters). The Shapley value of component $i$ is:
\begin{equation}
\phi_i(\bm{\theta}) = \sum_{S \subseteq \mathcal{N} \setminus \{i\}} \frac{|S|! \, (|\mathcal{N}| - |S| - 1)!}{|\mathcal{N}|!} \left[v(S \cup \{i\}) - v(S)\right]
\end{equation}
where $v(S) = \mathbb{E}_{\bm{\theta}_{-S}}[\Delta J(\bm{\theta}_S, \bm{\theta}_{-S})]$ is the expected gap when components in $S$ are set to experimental values.
\end{definition}

Recent research has demonstrated SHAP's effectiveness in explaining RL and robotic systems. \cite{beechey2023explaining} developed a theoretical framework applying Shapley values to RL value functions, addressing challenges of temporal and non-i.i.d.\ data in sequential decision-making. \cite{zhang2022explainable} applied SHAP to deep RL for power system emergency control, while \cite{engelhardt2024exploring} investigated SHAP's reliability in RL with multidimensional action spaces. \cite{onyekpe2022explainable} applied SHAP to enhance transparency in safety-critical autonomous vehicle positioning, \cite{remman2021robotic} used SHAP to explain DDPG policies for lever manipulation, and \cite{zhang2025shapley} developed SVMM using Shapley values for multimodal contributions in continuous control. These works highlight the growing adoption of SHAP for interpretable and trustworthy RL-based systems across diverse domains. However, the impact of RL configurations on generalizability has not been quantitatively explored, representing a gap that our work addresses by leveraging SHAP to analyze configuration impacts and provide actionable insights for improving generalizability in robotics.

%% file: sections/3_methodology.tex
\section{Methodology}
In this section, we first present a theoretical foundation that demonstrates the benefits of appropriate RL configurations in reducing the generalization gap, underscoring the importance of understanding the impact of configuration on model generalizability (Section~\ref{sec:theoretical_foundation}). Subsequently, we introduce a SHAP-based framework for configuration sampling, model training and evaluation, SHAP explainer construction, result analysis, and configuration selection guidance (Section~\ref{sec:configuration_framework}). This comprehensive methodology will robustly demonstrate its value in understanding and improving RL generalizability both theoretically and empirically.

\subsection{Theoretical Foundation}
\label{sec:theoretical_foundation}

We present two theorems: first, that the generalization bound depends on configuration sensitivity $\mathcal{S}(\bm{\theta})$, showing the gap is controllable via $\bm{\theta}$; second, that Shapley values decompose this gap into interpretable contributions, enabling us to understand $\mathcal{S}(\bm{\theta})$ and to seek $\bm{\theta}^*$ for minimal generalization gap.

\begin{theorem}[Generalization Bound]
\label{thm:bound}
The generalization gap in Definition~\ref{def:generalizability} can be bounded by:
\begin{equation}
\Delta J(\theta) \leq L(\pi_\theta) \cdot \|\omega_S - \omega_T\| + \epsilon_{\text{opt}}(\theta)
\end{equation}
where:
\begin{itemize}
    \item $L(\pi_\theta)$ is the configuration-dependent Lipschitz constant, which can be approximated by the sensitivity coefficient $S(\theta) := \left\|\nabla_\omega J(\pi_\theta; \omega)\big|_{\omega=\omega_S}\right\|$
    \item $\epsilon_{\text{opt}}(\theta)$ captures the optimization error, measuring the suboptimality of policy $\pi_\theta$ compared to the optimal policy $\pi^*_S$ in the source domain
\end{itemize}
\end{theorem}

This theorem is supported by the smoothness assumption of physics dependence stated in Equation~\ref{eq:lipschitz}, which ensures that small changes in physics parameters do not cause discontinuous performance jumps. The connection between the Lipschitz constant $L(\pi_\theta)$ and sensitivity $S(\theta)$ is established through Taylor expansion in Equation~\ref{eq:taylor_expansion}. In Pattern 3 (Section~\ref{sec:results}), we empirically demonstrate that generalization correlates with the factors identified in the bound.

For a fixed policy $\pi$, the expected return $J(\pi; \omega)$ is $L(\pi)$-Lipschitz continuous with respect to physics parameters $\omega$:
\begin{equation}
\label{eq:lipschitz}
|J(\pi; \omega_S) - J(\pi; \omega_T)| \leq L(\pi) \cdot \|\omega_S - \omega_T\|
\end{equation}

For smooth performance functions, we can apply first-order Taylor expansion around the source domain physics parameters $\omega_S$:
\begin{equation}
\label{eq:taylor_expansion}
\begin{gathered}
J(\pi_\theta; \omega_T) \approx J(\pi_\theta; \omega_S) + \nabla_\omega J(\pi_\theta; \omega)\big|_{\omega=\omega_S} \cdot (\omega_T - \omega_S) \\[0.5em]
\Rightarrow \quad |J(\pi_\theta; \omega_S) - J(\pi_\theta; \omega_T)| 
\leq \left\|\nabla_\omega J(\pi_\theta; \omega)\big|_{\omega=\omega_S}\right\| \cdot \|\omega_S - \omega_T\|
\end{gathered}
\end{equation}
which establishes the connection between the Lipschitz constant and the gradient norm for sensitivity: $L(\pi_\theta) \approx S(\theta) = \|\nabla_\omega J(\pi_\theta; \omega)|_{\omega=\omega_S}\|$.

\textbf{Key Insight:} Different configurations $\theta$ yield policies with varying sensitivities $S(\theta)$ to physics changes. Lower $S(\theta)$ indicates greater robustness to domain shift, contributing to a smaller generalization gap.

\begin{theorem}[SHAP-Based Generalization Optimization]
\label{thm:shapley_generalization}
Let $\phi_i(\bm{\theta})$ denote the Shapley value of component $i$ for configuration $\bm{\theta}$, measuring its marginal contribution to the generalization gap. Then:

\begin{enumerate}
    \item \textbf{Efficiency Property:} The Shapley values satisfy the efficiency axiom, completely decomposing the generalization gap:
    \begin{equation}
    \sum_{i=0}^{m} \phi_i(\bm{\theta}) = \Delta J(\bm{\theta}) - \Delta J(\bm{\theta}^{\text{baseline}})
    \end{equation}
    where $\Delta J(\bm{\theta}) = J_S(\pi_{\bm{\theta}}) - J_T(\pi_{\bm{\theta}})$ is the generalization gap.
    
    \item \textbf{Sensitivity Encoding:} Each Shapley value encodes how component $i$ affects domain sensitivity:
    \begin{equation}
    \phi_i(\bm{\theta}) \approx \frac{\partial \mathcal{S}(\bm{\theta})}{\partial \theta_i} \cdot (\theta_i - \theta_i^{\text{baseline}}) \cdot \|\bm{\omega}_S - \bm{\omega}_T\|
    \end{equation}
    where $\mathcal{S}(\bm{\theta})$ is the configuration's sensitivity to domain shift and $\|\bm{\omega}_S - \bm{\omega}_T\|$ measures domain divergence.
    
    \item \textbf{Optimization Principle:} The optimal configuration for generalization is obtained by:
    \begin{equation}
    \bm{\theta}^* = \argmin_{\bm{\theta}} \sum_{i=0}^{m} \phi_i(\bm{\theta}) \approx \argmin_{\bm{\theta}} \Delta J(\bm{\theta})
    \end{equation}
    where minimizing the sum of Shapley values corresponds to minimizing the generalization gap.
\end{enumerate}
\end{theorem}

This theorem establishes a principled framework for explainable generalization optimization. By decomposing the generalization gap into interpretable component contributions using SHAP, we can: (1) identify how configurations affect generalization, (2) quantify each component's sensitivity to domain shift, and (3) optimize configurations to minimize the gap through explainability. This theorem is empirically validated by our results in Pattern 4 (Section~\ref{sec:results}).

\textbf{Key Insight:} By decomposing the generalization gap into interpretable configuration contributions, minimizing Shapley values guides selection toward more generalizable configurations across domains.

\subsection{Configuration Framework}
\label{sec:configuration_framework}
\begin{figure}[t]
   \begin{center}
\includegraphics[width=1.0\linewidth]{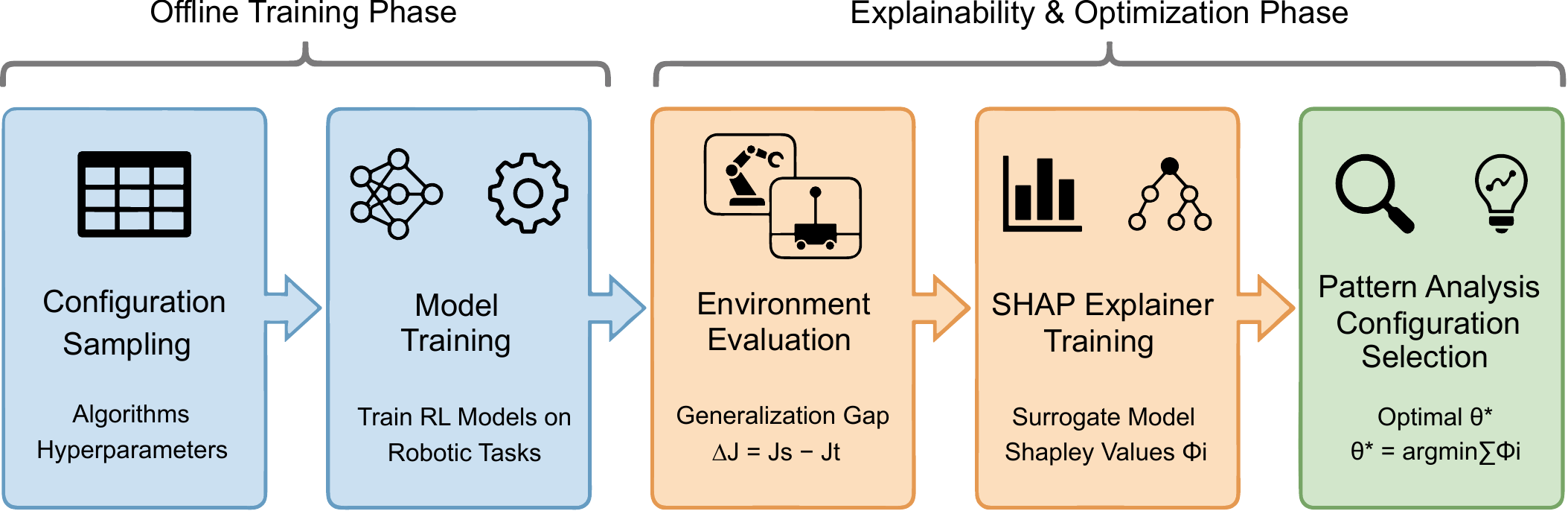}
   \end{center}
   \vspace{-0.2cm}
   \caption{SHAP-based configuration framework: (1) sample algorithm and hyperparameter configurations, (2) train RL models in source environment, (3) evaluate generalization gap across environments, (4) train surrogate SHAP explainer, and (5) analyze impact patterns and select optimal configurations. }
   \label{fig:shap_framework}
\end{figure}

We developed an explainable configuration framework that trains SHAP explainers and incorporates Shapley values to explain how RL configurations contribute to generalizability, as shown in Figure~\ref{fig:shap_framework}.

First, we define variable ranges for algorithms and hyperparameters and sample configurations, acknowledging that different algorithms utilize distinct hyperparameters. Second, we systematically train RL models using sampled configurations on robotic tasks in the training environment. Third, we evaluate model performance in both training and testing environments to calculate generalization gaps. Fourth, we apply surrogate explainability by training SHAP explainers that map configurations to generalization gaps. Finally, we compute feature impacts and derive insights, leading to learned sensitivity patterns and configuration selection that improve generalizability, as shown in Algorithm~\ref{alg:shap_optimization}.

\begin{algorithm}[h]
\caption{SHAP-Guided Configuration Selection}
\label{alg:shap_optimization}
\begin{algorithmic}[1]
\Require experimental data $\{(\bm{\theta}_j, J_S(j), J_T(j))\}_{j=1}^N$
\Ensure sensitivity $S(\bm{\theta})$, optimal configuration $\bm{\theta}^*$
\State compute generalization gaps: $\Delta J_j = J_S(j) - J_T(j)$ for $j=1,\ldots,N$
\State train surrogate model: $f(\bm{\theta}) \leftarrow \textsc{FitModel}(\{(\bm{\theta}_j, \Delta J_j)\}_{j=1}^N)$
\State compute Shapley values: $\phi_i(\bm{\theta}) \leftarrow \textsc{SHAP}_i(f, \bm{\theta})$ for $i=1,\ldots,m$
\State learn sensitivity model: $S(\bm{\theta}) \leftarrow \{\phi_i(\bm{\theta})\}_{i=1}^m$
\State optimize configuration: $\bm{\theta}^* = \argmin_{\bm{\theta}} \sum_{i=1}^{m} \phi_i(\bm{\theta})$
\State validate $S(\bm{\theta})$ and $\bm{\theta}^*$ per Theorem~\ref{thm:bound} and Theorem~\ref{thm:shapley_generalization}
\State \Return $S(\bm{\theta})$, $\bm{\theta}^*$
\end{algorithmic}
\end{algorithm}

%% file: sections/4_experiments.tex
\section{Experiments}
The experiments demonstrate the effectiveness of our configuration framework across multiple settings, where environments and tasks represent inherent variations that test generalizability, while algorithms and hyperparameters are the primary factors investigated. This section also presents implementation details, demonstrating the comprehensiveness, reliability, and reproducibility of our work.

\paragraph{Environments and Tasks. } We compare RL model performance by training and testing bidirectionally across two robotic simulation environments: \texttt{MuJoCo} and \texttt{PyBullet}~\cite{todorov2012mujoco,coumans2021pybullet}. These engines differ significantly in physics mechanisms, leading to distinct environmental dynamics $\|\omega_S - \omega_T\|$. MuJoCo employs constraint-based solvers with elliptic friction models, providing stable contact dynamics for continuous control~\cite{kaup2024review}, while PyBullet uses penalty-based methods that introduce greater variability in agent-environment interactions~\cite{yang2021open}. These fundamental differences make the two environments ideal for analyzing generalization across varying simulation dynamics. 

Standard locomotion tasks (\texttt{InvertedPendulum}, \texttt{HalfCheetah}, \texttt{Hopper}, \texttt{Walk er2D}) are chosen as their patterns serve as a baseline reference across varying complexity, suggesting the framework's scalability to higher-dimensional tasks in future work. These tasks are derived from OpenAI Gymnasium~\cite{towers2024gymnasium} and originally implemented in MuJoCo, with PyBullet alternatives~\cite{benelot2018} preserving identical observation spaces, action spaces, reward functions, and termination conditions, keeping the tasks comparable across environments despite physics differences.

\paragraph{Algorithms and Hyperparameters. } We implement four actor-critic algorithms (\texttt{PPO}, \texttt{A2C}, \texttt{DDPG}, \texttt{SAC}) from Stable-Baselines3~\cite{raffin2021stable-baselines3}, covering the dominant paradigm for continuous control in robotics. Actor-critic methods are most commonly used for these tasks, as value-based methods like DQN require discrete action spaces and model-based approaches add implementation complexity.  

\begin{table}[t]
\centering
\caption{Algorithm and hyperparameter configuration ranges}
\vspace{-0.2cm}
\begin{tabular*}{\textwidth}{@{\extracolsep{\fill}}llll}
\hline
\multicolumn{2}{l}{\textbf{PPO}}          & \multicolumn{2}{l}{\textbf{A2C}} \\ \hline
\texttt{learning\_rate}   & $(0.0001,0.01)$       & \texttt{learning\_rate}   & $(0.0001,0.01)$       \\
\texttt{gamma}            & $(0.8,0.999)$         & \texttt{gamma}            & $(0.8,0.999)$         \\
\texttt{clip\_range}      & $(0.1,0.3)$           & \texttt{gae\_lambda}      & $(0.9,0.99)$          \\
\texttt{n\_steps}         & $[128,512,1024,2048]$ & \texttt{vf\_coef}         & $(0.25,1.0)$          \\ \hline
\multicolumn{2}{l}{\textbf{DDPG}} & \multicolumn{2}{l}{\textbf{SAC}} \\ \hline
\texttt{learning\_rate}   & $(0.0001,0.01)$       & \texttt{learning\_rate}   & $(0.0001,0.01)$       \\
\texttt{gamma}            & $(0.8,0.999)$         & \texttt{gamma}            & $(0.8,0.999)$         \\
\texttt{tau}              & $(0.001,0.01)$        & \texttt{tau}              & $(0.001,0.01)$        \\
\texttt{buffer\_size}     & $[10^4, 10^5, 10^6]$ & \texttt{ent\_coef}   & $(0.01,0.2)$          \\ \hline
\end{tabular*}
\vspace{-0.2cm}
\label{tab:configurations}
\end{table}

These algorithms span on-policy versus off-policy learning, deterministic versus stochastic policies, and different exploration mechanisms, enabling systematic evaluation of algorithmic design choices on generalization. For each algorithm, we select four key hyperparameters: two shared (learning rate controlling update magnitudes, and gamma balancing immediate versus long-term rewards) and two algorithm-specific parameters corresponding to their underlying RL mechanisms. Detailed ranges and settings are in Table~\ref{tab:configurations}.


\paragraph{Implementation Details. } 
We sampled 100 configurations per algorithm per task bidirectionally, using 3 random seeds per configuration for robustness (9,600 total models). Learning rate and gamma were sampled from log-uniform distributions, while other hyperparameters used uniform distributions. Algorithms were one-hot encoded from on-policy to off-policy: (0) PPO, (1) A2C, (2) DDPG, and (3) SAC. Each model was trained for 100,000 steps in one environment and tested in the other, with generalization gap measured as $\Delta J(\theta) = J_S(\pi_\theta) - J_T(\pi_\theta)$ per Definition~\ref{def:generalizability}, preserving directional transfer information.

Experiments were conducted on GPU servers with 1GB usage per thread, totaling 864 training hours across multiple threads. On-policy algorithms (PPO, A2C) required $\sim$96 hours each, while off-policy algorithms (DDPG, SAC) required 288 and 384 hours respectively due to higher sample complexity. Post-training SHAP analysis was lightweight, requiring only minutes per case, with the modular framework enabling straightforward adaptation to other domains through simple environment and task modifications.



%% file: sections/5_results.tex
\section{Results and Discussion} \label{sec:results}
The results focus on four main patterns: (1) main impacts of RL algorithms and hyperparameters, (2) interaction impacts between configurations, (3) task and environment insights empirically validating Theorem~\ref{thm:bound} for the correlation between generalizability and configurations, and (4) configuration selection empirically validating Theorem~\ref{thm:shapley_generalization} through SHAP-guided generalizability optimization.

\paragraph{Pattern 1: Main Impacts. }
We employ beeswarm plots (Figure~\ref{fig:exp1}) to visualize Shapley values and analyze how hyperparameters impact generalization across algorithms. Since the generalization gap is $\Delta J(\theta) = J_S(\pi_\theta) - J_T(\pi_\theta)$, lower Shapley values indicate configurations that improve generalizability. The x-axis range is standardized ($-500$ to $+1250$) across all plots to facilitate comparison, revealing distinct patterns across the four algorithms:

\begin{figure}[t]
   \begin{center}
\includegraphics[width=1.0\linewidth]{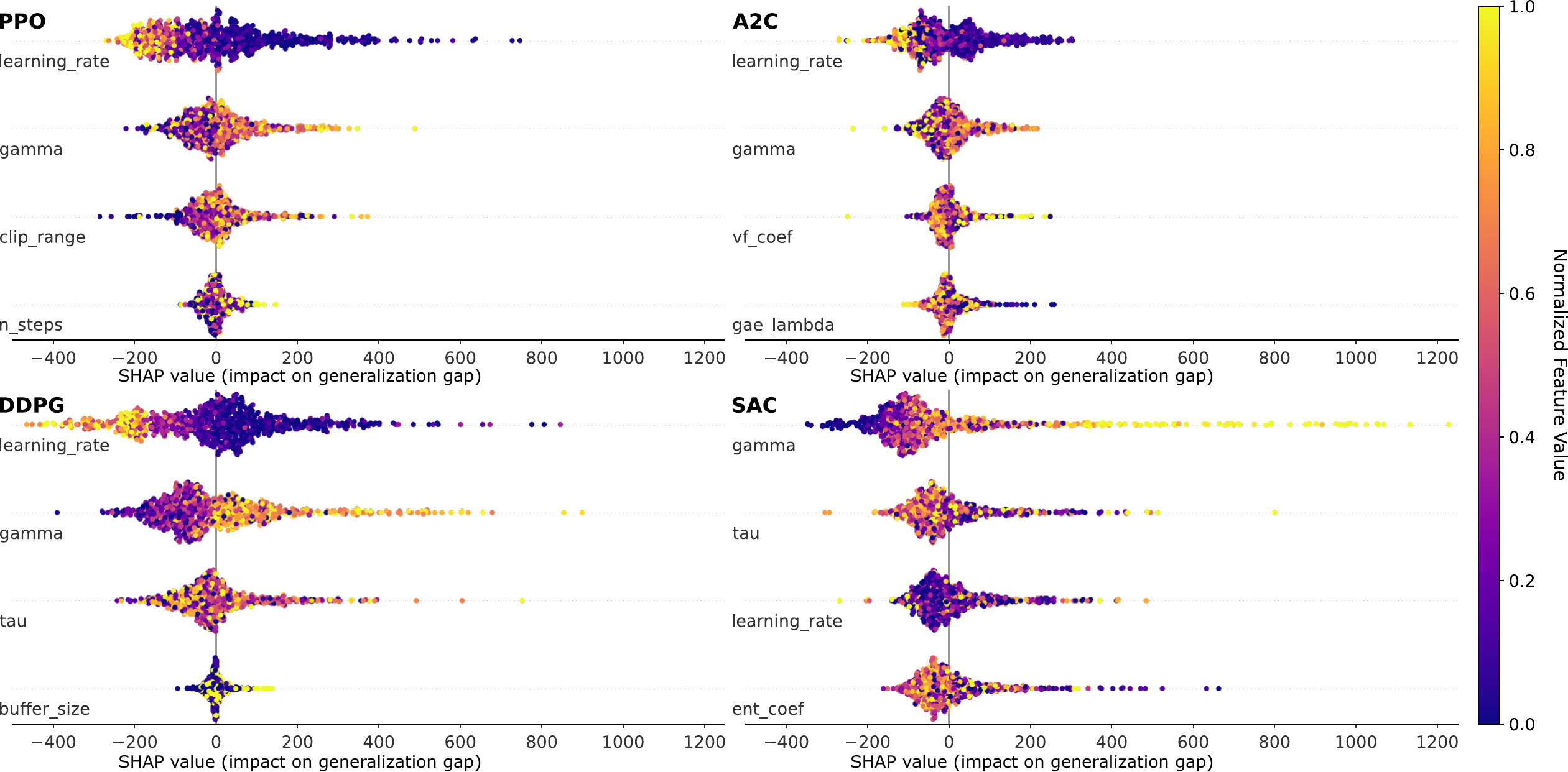}
   \end{center}
   \vspace{-0.2cm}
   \caption{Main impact patterns across all algorithms and hyperparameters, lower (leftward) SHAP values indicate better generalizability. }
   \vspace{-0.2cm}
   \label{fig:exp1}
\end{figure}

\begin{itemize}[leftmargin=*]
    \item In \texttt{PPO}, learning rate is most influential, with higher values promoting better generalization through faster adaptation. Higher gamma leads to overfitting to training-specific reward patterns, while larger clip ranges encode training-specific behaviors, both resulting in elevated Shapley values. The n\_steps parameter shows weak and interaction-dependent impact.
    \item \texttt{A2C} exhibits the lowest hyperparameter sensitivity, likely due to its synchronous update mechanism that regularizes learning. Higher learning rates and lower gamma improve generalizability, with the latter reducing the temporal horizon and preventing trajectory overfitting. The vf\_coef and gae\_lambda show bidirectional effects dependent on environment-specific factors.
    \item \texttt{DDPG} exhibits strong hyperparameter sensitivity. Lower learning rates hurt generalizability as the deterministic policy overfits to training-specific trajectories, while lower gamma improves it by smoothing environment-specific noise through long-term value estimation. Tau and buffer\_size patterns depend on the exploration-exploitation balance.
    \item \texttt{SAC} is an exception where learning rate ranks third, as entropy regularization already constrains policy updates. Gamma is most critical, with lower values reducing gaps. Tau and ent\_coef exhibit ambiguous effects.
\end{itemize}

The varying hyperparameter sensitivity reflects fundamental algorithmic differences. On-policy methods (PPO, A2C) show clearer learning rate patterns as they update directly from current experience, making learning speed critical for generalization. Off-policy methods (DDPG, SAC) exhibit more complex patterns with larger SHAP variations due to replay mechanisms and target network interactions. Learning rate and gamma show consistent patterns across algorithms (except SAC), while algorithm-specific hyperparameters have a relatively ambiguous impact. These findings provide actionable guidance for algorithm and hyperparameter selection when prioritizing generalizability.

\begin{figure}[t]
   \begin{center}
\includegraphics[width=1.0\linewidth]{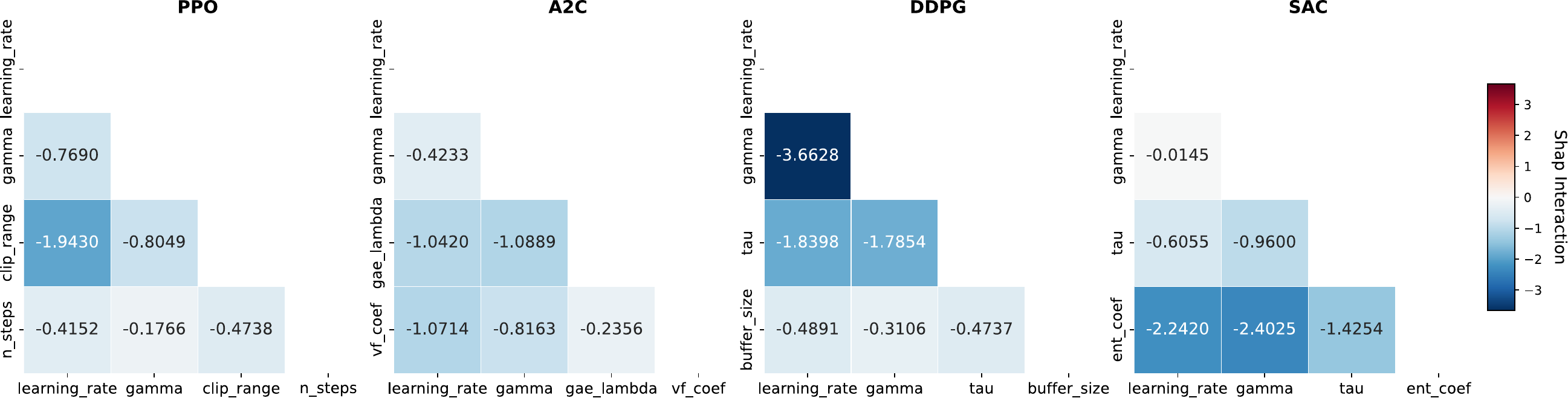}
   \end{center}
   \vspace{-0.2cm}
   \caption{Feature interactions between hyperparameters in all four algorithms, where darker blue indicates stronger beneficial interactions. }
   \vspace{-0.2cm}
   \label{fig:exp2}
\end{figure}

\paragraph{Pattern 2: Interaction Impacts. }
\begin{wrapfigure}[15]{r}{0.5\textwidth}  
   \centering
   \vspace{-0.8cm}
   \includegraphics[width=0.48\textwidth]{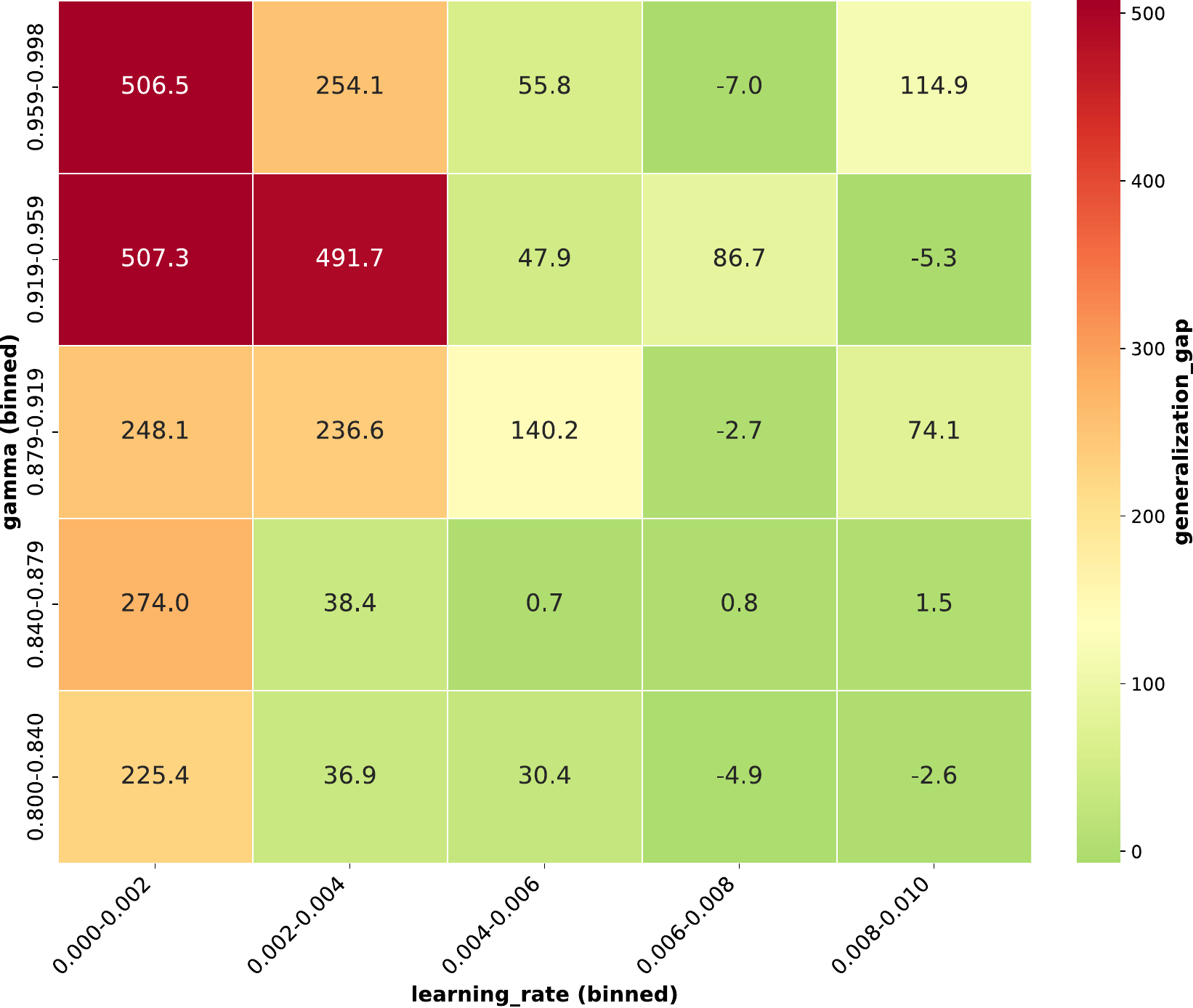}
   \vspace{-0.1cm}
   \caption{Feature interaction between learning rate and gamma in DDPG algorithm. }
   \label{fig:exp22}
\end{wrapfigure}

RL algorithms and hyperparameters not only have individual impacts on generalizability but also interact in complex ways. As shown in Figure~\ref{fig:exp2}, Shapley interaction values quantify these interdependencies by measuring $\phi_{i,j} = \Delta_{ij} - \Delta_i - \Delta_j$, where $\Delta_{ij}$ represents the joint marginal contribution of features $i$ and $j$, and $\Delta_i$, $\Delta_j$ their individual contributions, with lower values (darker blue) indicating better generalizability. Notably, the learning\_rate vs. gamma interaction is insignificant in PPO, A2C, and SAC but highly impactful in DDPG ($-3.6628$). Other strong beneficial interactions include learning\_rate vs.\ clip\_range ($-1.9430$) in PPO, and learning\_rate vs. ent\_coef ($-2.2420$) and gamma vs. ent\_coef ($-2.4025$) in SAC.

We specifically examine the \texttt{learning\_rate vs. gamma} interaction in DDPG in Figure~\ref{fig:exp22}, where lower (greener) values indicate better generalization. Higher learning rates ($0.006$-$0.01$) with lower gamma ($0.800$-$0.879$) achieve consistently good generalization with near-zero gaps (e.g., $-4.9$, $-2.6$), while lower learning rates ($0.0001$-$0.004$) with higher gamma ($0.919$-$0.999$) severely degrade generalizability with gaps exceeding $400$-$500$, explaining the substantial interaction importance ($-3.6628$). Higher learning rates enable faster adaptation, while lower gamma prevents overfitting to long-term training dynamics, whereas slow learning combined with high gamma causes overfitting to specific trajectory patterns, resulting in poor transfer.

\begin{figure}[t!]
   \begin{center}
\includegraphics[width=1.0\linewidth]{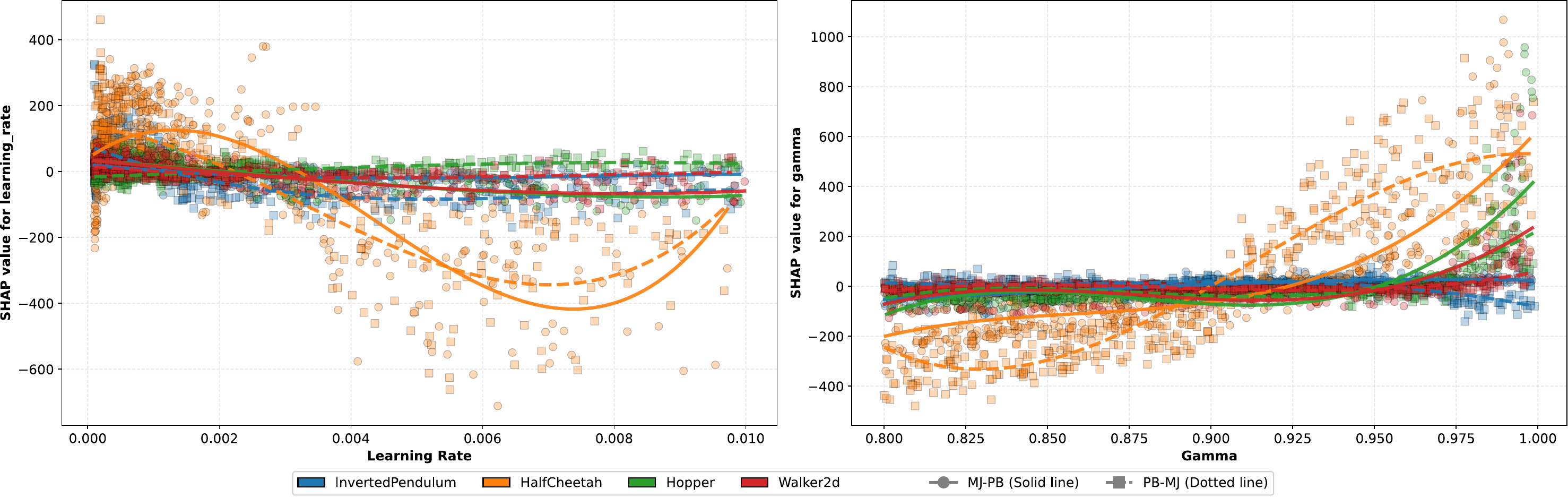}
   \end{center}
   \vspace{-0.2cm}
   \caption{SHAP dependence of learning rate and gamma across tasks and environments. }
   \vspace{-0.2cm}
   \label{fig:exp3}
\end{figure}

\paragraph{Pattern 3: Task and Environment Insights. }
In Figure~\ref{fig:exp3}, we examine results across four tasks and two transfer directions: MJ-PB (trained on MuJoCo, tested on PyBullet) and PB-MJ (reverse). We focus on learning rate and gamma as the most impactful shared hyperparameters to understand the environmental difference $\|\omega_S - \omega_T\|$ and Lipschitz constant $L(\pi_\theta)$. Different tasks show distinct sensitivity patterns. \texttt{HalfCheetah} exhibits the highest sensitivity to hyperparameters, with large Shapley value variations and pronounced U-shaped curves, while simpler tasks like \texttt{InvertedPendulum} remain stable across most ranges. This correlates with task complexity and stability: simpler tasks are less affected by configuration selection. Notably, bidirectional transfers (\texttt{MJ-PB vs. PB-MJ}) show similar sensitivity patterns with overlapping polynomial trends. This suggests the environmental difference $\|\omega_S - \omega_T\|$ is symmetric between the two simulators, as both transfer directions involve the same pair of environments. However, policy sensitivity $L(\pi_\theta)$ still varies and dominates the generalization gap due to differences in the training environment. These findings empirically support Theorem~\ref{thm:bound}, confirming that configuration sensitivity and environmental difference are the primary factors affecting generalization performance.



\paragraph{Pattern 4: Configuration Selection Validation. }

\begin{figure}[t!]
   \begin{center}
\includegraphics[width=1.0\linewidth]{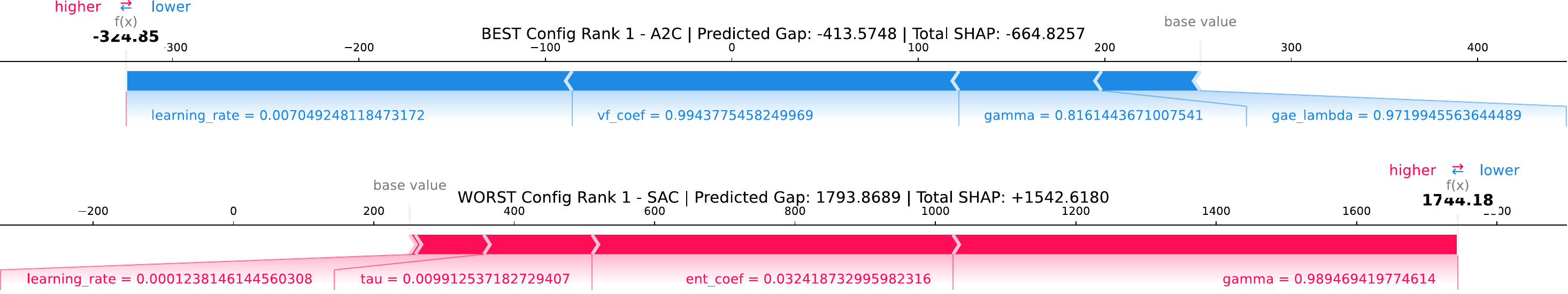}
   \end{center}
   \vspace{-0.2cm}
   \caption{Best and worst configurations identified by SHAP analysis. }
   \label{fig:exp4}
   \vspace{-0.2cm}
\end{figure}

Having established the SHAP-guided selection strategy, we use the SHAP explainer to predict the optimal and worst configurations. As shown in Figure~\ref{fig:exp4}, the predicted results validate our understanding of the patterns. The optimal \texttt{A2C} configuration features a high learning rate (0.0070), high vf\_coef (0.9943), low gamma (0.8161), and moderate gae\_lambda (0.9719), aligning with our understanding that higher learning rates and lower gamma values improve generalizability, achieving a predicted gap of $-413.5748$. Conversely, the worst configuration features \texttt{SAC} with high gamma (0.9894), low learning rate (0.0001), moderate ent\_coef (0.0324), and high tau (0.0099), exactly matching our understanding of SAC configurations that contribute to poor generalizability, with a predicted gap of $1793.8689$. Together, these results demonstrate that the SHAP explainer reliably captures the system's generalization trends.

\begin{table}[ht]
\centering
\caption{Closest matching configurations to SHAP-predicted and worst (hp1: learning\_rate, hp2: gamma, hp3: gae\_lambda / tau, hp4: vf\_coef / ent\_coef.). }
\label{tab:validation_configs}
\setlength{\tabcolsep}{5pt}
\begin{tabularx}{\linewidth}{@{\extracolsep{\fill}}l l l l l l l}
\toprule
\textbf{Task} & \textbf{Transfer} &
\textbf{hp1} & \textbf{hp2} & \textbf{hp3} & \textbf{hp4} &
\textbf{$\Delta J$} \\
\midrule
\multicolumn{7}{l}{\small A2C best configuration (predicted gap: $-413.57$;\; actual average: $-132.58$)} \\
\midrule
HalfCheetah     & MJ$\to$PB & 0.0071 & 0.8269 & 0.9093 & 0.9937 & $-815.73$ \\
HalfCheetah     & PB$\to$MJ & 0.0079 & 0.8397 & 0.9457 & 0.7981 & $-45.12$  \\
Hopper          & MJ$\to$PB & 0.0066 & 0.8104 & 0.9807 & 0.7580 & $+45.65$  \\
Hopper          & PB$\to$MJ & 0.0062 & 0.8605 & 0.9606 & 0.8326 & $-10.14$  \\
InvertedPendulum& MJ$\to$PB & 0.0077 & 0.8589 & 0.9505 & 0.9852 & $-267.80$ \\
InvertedPendulum& PB$\to$MJ & 0.0069 & 0.8159 & 0.9444 & 0.7739 & $+103.17$ \\
Walker2D        & MJ$\to$PB & 0.0098 & 0.8066 & 0.9538 & 0.8200 & $-21.48$  \\
Walker2D        & PB$\to$MJ & 0.0062 & 0.8119 & 0.9221 & 0.7109 & $-49.15$  \\
\midrule
\multicolumn{7}{l}{\small SAC worst configuration (predicted gap: $+1793.87$;\; actual average: $+1223.56$)} \\
\midrule
HalfCheetah     & MJ$\to$PB & 0.0014 & 0.9853 & 0.0088 & 0.0284 & $+5174.93$ \\
HalfCheetah     & PB$\to$MJ & 0.0001 & 0.9667 & 0.0094 & 0.0408 & $+1513.40$ \\
Hopper          & MJ$\to$PB & 0.0003 & 0.9508 & 0.0082 & 0.0305 & $+355.21$  \\
Hopper          & PB$\to$MJ & 0.0015 & 0.9917 & 0.0061 & 0.0257 & $+1039.13$ \\
InvertedPendulum& MJ$\to$PB & 0.0001 & 0.9727 & 0.0095 & 0.0314 & $+251.07$  \\
InvertedPendulum& PB$\to$MJ & 0.0002 & 0.9853 & 0.0035 & 0.0290 & $+261.57$  \\
Walker2D        & MJ$\to$PB & 0.0008 & 0.9911 & 0.0056 & 0.0368 & $+997.40$  \\
Walker2D        & PB$\to$MJ & 0.0001 & 0.9466 & 0.0071 & 0.0237 & $+195.77$  \\
\bottomrule
\end{tabularx}
\label{tab:results}
\end{table}

To further validate the SHAP explainer predictions, we locate real configurations with the closest matching hyperparameters to the predicted best and worst configurations across tasks and transfer directions, as these predicted configurations do not exist in the training dataset. Table~\ref{tab:results} presents these closest matches and their corresponding generalization behavior. The optimal \texttt{A2C} configuration consistently achieves low generalization gaps (6 of 8 cases negative, actual average: -132.58, predicted: -413.57), while the worst \texttt{SAC} configuration exhibits poor generalizability in all cases (actual average: 1223.56, predicted: 1793.87). The directional prediction is correct in all cases, and although absolute gap values vary across tasks, the consistent generalization trends validate that SHAP-guided configuration selection provides meaningful and proactive guidance for optimizing generalizability, supporting Theorem~\ref{thm:shapley_generalization}.


%% file: sections/6_conclusion.tex
\section{Conclusion}


In conclusion, our work addresses configuration pattern analysis for transfer generalizability in RL through explainable AI. We establish theoretical foundations showing that generalizability can be systematically optimized via SHAP analysis, and propose a modular framework validated across robotic tasks in MuJoCo and PyBullet environments. Our analysis reveals key insights into configuration patterns, feature interactions, and task-specific effects, providing practitioners with actionable guidance for intelligently selecting configurations to enhance generalizability in Sim2Sim and Sim2Real transfer scenarios. We envision this work as a general foundation for broader generalizability optimization across configurations, tasks and environments in RL for robotics.